\documentclass[10pt,twocolumn,letterpaper]{article}

\usepackage{ijcb}
\usepackage{times}
\usepackage{epsfig}
\usepackage{graphicx}
\usepackage{amsmath}
\usepackage{amssymb}
\usepackage{array}
\usepackage{xspace}
\usepackage[caption=false]{subfig}
\usepackage{paralist}
\usepackage{url}
\usepackage{multirow}
\usepackage[norule,symbol,perpage]{footmisc}

\usepackage[pagebackref=true,breaklinks=true,letterpaper=true,colorlinks,bookmarks=false]{hyperref}

\ijcbfinalcopy 

\ifijcbfinal\fi

\makeatletter
\def\ps@IEEEtitlepagestyle{
\def\@oddfoot{\mycopyrightnotice}
\def\@evenfoot{}
}
\def\mycopyrightnotice{
{\hfill \footnotesize 978-1-7281-9186-7/20/\$31.00 \copyright 2020 IEEE\hfill}
}
\makeatother

\begin{document}

\title{Cross-Domain Identification for Thermal-to-Visible Face Recognition}

\author{Cedric Nimpa Fondje\renewcommand*{\thefootnote}{\arabic{footnote}}\footnotemark[1]$\;^,$\renewcommand*{\thefootnote}{\fnsymbol{footnote}}\footnotemark[1]\\
\and
\renewcommand*{\thefootnote}{\arabic{footnote}}
Shuowen Hu\footnotemark[2]\\
\and
\renewcommand*{\thefootnote}{\arabic{footnote}}
Nathaniel J. Short\footnotemark[2]$\;^,$\footnotemark[3]\\
\and
Benjamin S. Riggan\renewcommand*{\thefootnote}{\arabic{footnote}}\footnotemark[1]$\;^,$\renewcommand*{\thefootnote}{\fnsymbol{footnote}}\footnotemark[1]\\
\and
\renewcommand*{\thefootnote}{\arabic{footnote}}\footnotemark[1]~~University of Nebraska-Lincoln, 1400 R St, Lincoln, NE 68588\\
\renewcommand*{\thefootnote}{\arabic{footnote}}\footnotemark[2]~~CCDC Army Research Laboratory, 2800 Powder Mill Rd., Adelphi, MD 20783\\
\renewcommand*{\thefootnote}{\arabic{footnote}}\footnotemark[3]~~Booz Allen Hamilton, 8283 Grennsboro Dr., McLean, VA 22102
\and
\footnotemark[1]~~\emph{Corresponding authors: cedricnimpa@huskers.unl.edu, briggan2@unl.edu}
}

\maketitle
\thispagestyle{empty}

\begin{abstract}
   Recent advances in domain adaptation, especially those applied to heterogeneous facial recognition, typically rely upon restrictive Euclidean loss functions (e.g., $L_2$ norm) which perform best when images from two different domains (e.g., visible and thermal) are co-registered and temporally synchronized. 
   This paper proposes a novel 
   domain adaptation framework that combines a new feature mapping sub-network with existing deep feature models, which are based on modified network architectures (e.g., VGG16 or Resnet50).  This framework is optimized by introducing new cross-domain identity and domain invariance loss functions for thermal-to-visible face recognition, which alleviates the requirement for precisely co-registered and synchronized imagery.
   We provide extensive analysis of both features and loss functions used, and compare the proposed domain adaptation framework with state-of-the-art feature based domain adaptation models on a difficult dataset containing facial imagery collected at varying ranges, poses, and expressions. Moreover, we analyze the viability of the proposed framework for more challenging tasks, such as non-frontal thermal-to-visible face recognition.
   
\end{abstract}


\section{Introduction}

Facial recognition (FR) systems have become more ubiquitous, being deployed across diverse services, including popular social media platforms, consumer devices (e.g., smart phones), and local/federal government or law enforcement databases (e.g., DoD Automated Biometric Identification System~\cite{ABISreport}).
The algorithms/models in these FR systems are almost exclusively developed for matching visible spectrum facial imagery, due to the ubiquity of low-cost visible cameras with increasingly high resolution. Advances in visible spectrum FR algorithms, such as enhanced robustness to a varying pose, illumination, expression, resolution, and partial occlusion conditions, can be partially attributed to the proliferation of large-scale FR datasets (e.g. LFW ~\cite{LFWTech}, MegaFace Challenge 1 ~\cite{KemelmacherShlizerman2016TheMB}, and MegaFace Challenge 2 ~\cite{8099846}), computational resources (e.g., GPUs ~\cite{KemelmacherShlizerman2016TheMB}), and deep learning models (e.g., convolutional neural networks or generative adversarial networks). 

However, there has been growing interest in heterogeneous facial recognition (HFR), such as matching facial signatures in near infrared (NIR)~\cite{10.1007/978-3-540-74549-5_55, 5597000, 6595898} and thermal infrared~\cite{Hu:15, Gurton:14, Short:15} 
images to visible facial signatures, to better facilitate FR under low-light and variable illumination settings (e.g., nighttime FR). Unlike within spectrum matching (e.g., visible-to-visible FR), HFR---especially thermal-to-visible FR---has primarily been studied under significantly limited conditions (e.g., frontal images) due to the increased cost and complexity necessary to collect datasets containing visible and thermal face imagery that is on par with the size and number of conditions consistent with visible FR benchmarks.  
     
Therefore, in this paper, we facilitate the advancement of thermal-to-visible FR research by introducing a new state-of-the-art domain adaptation framework and demonstrate its robustness on a difficult multi-modal face dataset, which incorporates more subjects and conditions than previous datasets.
Since existing databases/watch-lists across academia, industry, government, and law enforcement currently enroll only visible face imagery, matching thermal face images with visible face imagery is necessary to perform FR in settings with low or variable light, which is significantly more difficult than conventional visible-to-visible FR in practice. This increased difficulty arises from significant differences between the thermal and visible face signatures~\cite{10.1117/12.920330}. 

Recently, several domain adaptive machine learning techniques have been used to reduce the modality gap between visible and thermal imagery. Hu et al.~\cite{Hu:15} demonstrated that using thermal cross-examples in a one-vs-all framework using partial least squares (PLS) classifiers enhanced discriminability when matching histogram of oriented gradient (HOG) features between thermal and visible images.  Sarfraz et al.~\cite{DBLP:conf/bmvc/SarfrazS15} addressed the problem of thermal-to-visible FR using a small (shallow) neural network to perform a direct regression between local features extracted from thermal images and corresponding visible features that showed promising results, especially given the limited data used to train the network. Riggan et al.~\cite{7270978} used a coupled auto-associative model to learn a discriminative common latent subspace between visible and thermal image patches, and later~\cite{7477447} introduced an optimal feature regression and discriminative framework that exploited a combination of elements from ~\cite{Hu:15},~\cite{DBLP:conf/bmvc/SarfrazS15}, and~\cite{7270978}. More recently, there have been domain adaptive techniques, such as~\cite{8411218,8987416} that utilize a pre-trained deep learning model and domain adaptive convolutional neural networks (CNNs) to learn more global (contextual) domain-invariant features for thermal-to-visible FR.  However, these techniques rely upon polarimetric thermal imagery and a relatively larger face crop size and are more sensitive to pose variations and temporal changes to the face (e.g., hair style, facial hair, aging).
      
The primary objective of this work is to enhance thermal-to-visible FR performance. 
Our contributions include:
\begin{compactitem}
    \item modified VGG16 \cite{Simonyan15} and Resnet50 \cite{7780459} architectures for feature extraction, 
    \item a new feature mapping sub-network to help bridge the domain gap,
    \item a new cross-domain identification loss function to relax requirement for precise co-registration and synchronization,
    \item a new domain invariance loss function provide a type of cross-domain regularization. 
\end{compactitem}
Compared to state-of-the-art feature-based domain adaptation methods~\cite{DBLP:conf/bmvc/SarfrazS15, 7477447, DBLP:journals/ijcv/SarfrazS17}, the proposed framework (Section~\ref{sec:method}) achieves enhanced conventional thermal-to-visible FR performance using an expanded multi-modal face dataset from the U.S.~Army CCDC Army Research Laboratory~\cite{article}, which contains frontal imagery (visible and polarimetric thermal) with neutral (baseline) and variable expressions.
Moreover, we present new state-of-art results on a dataset that contains significantly more visible and thermal image pairs from a larger population of subjects, which is evaluated using imagery under varying pose and expression conditions.  We intentionally exclude polarimetric thermal face signatures from our evaluations, since polarimetric thermal imaging is still an emerging area, whereas conventional thermal imagers have been widely deployed for military and homeland security applications, and is even becoming more widely available for commercial applications.



\section{Heterogeneous Face Recognition}

In this section, HFR models that have demonstrated some success are briefly reviewed so that we may compare and contrast these approaches with our proposed methodology (Section~\ref{sec:method}).  In particular, we discuss two general approaches to HFR: transfer learning and domain adaptation.

\subsection{Transfer Learning}

Goodfellow et al.~\cite{Goodfellow-et-al-2016} describe transfer learning as a situation where knowledge learned in one setting is exploited to improve generalization in another setting. Often, 
the knowledge (features, weights, etc.) is transferred from previously trained models for one task to modified models for a different task in order to mitigate issues with over-fitting on the new task. 

Intuitively, it seems that some shared information exists between corresponding visible and thermal face images, but the optimal form of knowledge transfer is not obvious.  Instance, feature, parameter, and relational knowledge are four different types of knowledge that may be transferred between domains~\cite{5288526}, where feature and
parameter transfer learning are the most common.

The simplest form of parameter transfer learning includes weight sharing for deep neural networks, where usually the low-level parameters are assumed to be more applicable to both domains and are shared between two models.  However, the high-level, more application specific, features are less generalizable and are fine-tuned to the desired task. More complex forms of parameter transfer learning, such as enforcing parameters to be related via a linear transform~\cite{8310033}, may also be used. However, when domain gaps are very large (as with thermal and visible images), parameter constraints may be too restrictive to effectively learn a common representation for cross-domain FR.  Therefore, we use domain adaptation. 

\subsection{Domain Adaptation}

Unlike transfer learning, domain adaptation assumes a large shift between the source (image) distributions.  The large perceptual gap between thermal and visible images is due to the fact that visible images are acquired using “reflected” light off of objects/faces and thermal images are acquired using ``emitted'' thermal radiation emanating from objects/faces. Sharing low-level weights may not be optimal since visible and thermal facial imagery exhibit a highly non-linear relationship and contain disparate information---thermal imagery has less high frequency and geometric details compared to visible imagery.

\subsubsection{Deep Perceptual Mapping}
Sarfraz et al.~\cite{DBLP:conf/bmvc/SarfrazS15} introduced the Deep Perceptual Mapping (DPM) specifically for thermal-to-visible FR.
With DPM, thermal-to-visible FR is assimilated as a regression problem, where a multilayer neural network directly regresses features (e.g., DSIFT or HOG) by minimizing 
\begin{equation}
J_{dpm}(\mathbf{\Theta})=\sum \left\|\mathbf{y}-f_{dpm}(\mathbf{x};\mathbf{\Theta}) \right \|^2,
\label{eq:dpm}
\end{equation}
where $\mathbf{y}$ denotes a thermal feature vector, $f_{dpm}(\mathbf{x};\mathbf{\Theta})$ denotes the DPM estimate for the thermal feature vector from a given visible feature vector ($x$), and $\mathbf{\Theta}$ is the set of trainable model parameters.

In ~\cite{DBLP:conf/bmvc/SarfrazS15}, a three-layer DPM is optimized to predict thermal DSIFT features given the corresponding DSIFT features from the visible domain. Specifically, the authors show that extracting DSIFT features followed by principal components analysis (PCA) dimensionality reduction was effective for thermal-to-visible FR on the UND X1 database~\cite{UNDX1}.  Image representations are constructed by concatenating local feature vectors extracted from overlapping image patches. A gallery is constructed from the estimated thermal representations of the visible gallery images, and matching is performed by computing the cosine similarity between the actual image representation from a thermal probe image and the predicted thermal feature representations from the gallery.

Similarly, our framework incorporates some compression in order to extract the most representative information from visible and thermal feature embedding representations.  However, instead of DSIFT features, we integrate features extracted from deep neural networks that exhibit an effective number of features (i.e., channels), receptive fields size, and non-linearity, while also alleviating potential for over-fitting. 

\subsubsection{Coupled Neural Networks (CpNN) }
An alternative approach for thermal-to-visible FR is a coupled neural network (CpNN), which learns how to extract common latent features between visible and thermal imagery by optimizing
\begin{equation}
    J_{cpnn}(\mathbf{\Theta}_x,\mathbf{\Theta}_y)=\left\|f(\mathbf{x}, \mathbf{\Theta}_x) - g(\mathbf{y};\mathbf{\Theta}_x) \right\|^2,
\end{equation}
where $f(\mathbf{x}, \mathbf{\Theta}_x)$ and $g(\mathbf{y}, \mathbf{\Theta}_y)$ are mapped visible and thermal features, respectively.  Unlike DPM, this approach attempts to find two mappings such that features are sufficiently close in the mean square sense.  

CpNNs train two encoders---one for the visible domain and one for the thermal domain---to produce similar features for corresponding inputs. The primary difference between CpNNs and DPMs is that CpNNs learn how to extract similar features, rather than extracting features from each domain and learning a mapping explicitly between these features [13].  

Similar to ~\cite{7270978, 7477447}, ~\cite{8411218} uses coupled networks to perform polarimetric thermal-to-visible FR.  However, they use global average pooling with a VGG-like CNN architecture instead of local-features to form the common embedding representations.  The motivation for global average pooling is to remove parameter intensive fully connected layers, which can potentially lead to over-fitting.  However, one disadvantage is that global pooling assumes that the effective receptive field for the network is sufficiently large to provide enough contextual information to perform facial verification or identification, which requires additional layers and parameters. Moreover, in practice, it is generally easier to minimize the apparent modality gaps over relatively small image regions, but at the risk of needing more applications of local mappings to provide holistic image representation. 
Therefore, our proposed approach aims to learn local features with the largest context possible, but without the use of global pooling.  

\subsubsection{Generative Adversarial Network (GANs)}
Generative Adversarial Networks~\cite{10.5555/2969033.2969125} are composed of two parts: a generator ($G(\mathbf{z})$) and discriminator ($D(\mathbf{x})$), which are trained to minimize and maximize Eq.~\ref{eq:adversarialloss} with respect to $G$ and $D$, respectively.
\begin{equation}
    \mathbb{E}_{\mathbf{x}}[\log D(\mathbf{x})] + \mathbb{E}_{\mathbf{z}}[1-\log D(G(\mathbf{z}))].
    \label{eq:adversarialloss}
\end{equation}
The generator aims to confuse the discriminator by randomly synthesizing realistic samples from some underlying distribution.  In the context of HFR, instead of generating random, realistic faces from a random vector, $\mathbf{z}$, conditional GANs (CGANs) are used to synthesize specific images from domain corresponding to a condition input, i.e., an image from another domain.

Similar to Eq.~\ref{eq:adversarialloss}, CGANs optimize 
\begin{equation}
    \mathbb{E}_{\mathbf{x}}[\log D(\mathbf{x|y})] + \mathbb{E}_{\mathbf{z}}[1-\log D(G(\mathbf{z|y}))],
    \label{eq:conditionaladversarialloss}
\end{equation}
using the same minimax optimization used for GANs.  Specifically for HFR, much like image-to-image translation techniques~\cite{8100115, articleBEGAN} that perform image style transfer, \cite{article} generates high-quality visible face images from given polarimetric thermal face imagery.  The disadvantage to this type of approach is computational complexity of the model.  In~\cite{8551516}, a similar network reported an average run time of more than 270 seconds per image.  Therefore, such a computationally complex approach is not suitable for real-time FR, but may be a useful tool for analyst and less time critical activities.

Even GAN-based architectures, which use encoder-decoder networks for the generators, rely on robust, discriminative embedding representations.  For example, in addition to the adversarial loss (e.g., Eq.~\ref{eq:conditionaladversarialloss}), ~\cite{article} uses Euclidean loss functions between visible and thermal features to enforce similarity between feature embeddings.  This not only ensures photo-realism at the generators output, but also similar generated visible latent features from a given thermal image.  Therefore, we expect that our proposed approach could also be used as a complementary loss to provide additional guideance for GANs.  However, this is beyond the scope of this paper.


\begin{figure*}[htb]
\centering
\includegraphics[width=0.7\textwidth]{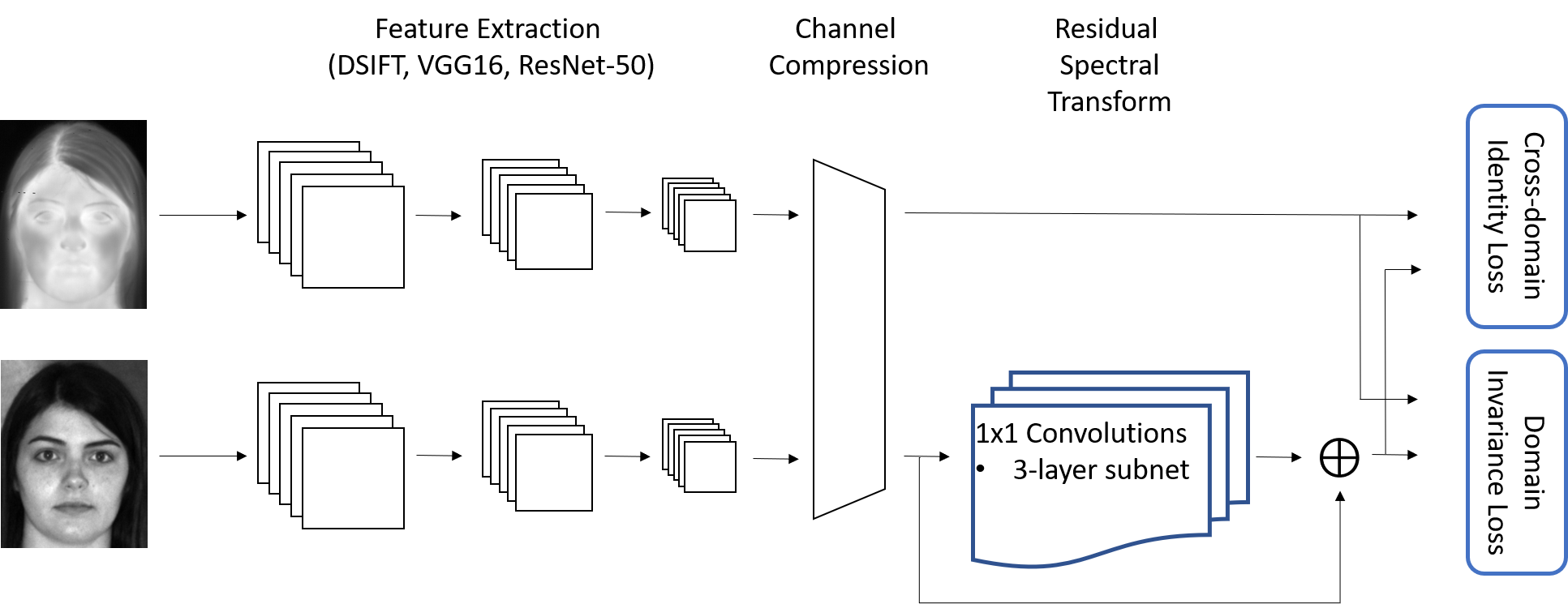}
\caption{Proposed domain adaptation model.}
\label{fig:model}
\end{figure*}

\section{Problem Definition}

Let $\mathbf{V}$ and $\mathbf{T}$ each be the set of $n$-dimensional descriptive feature vectors from the visible domain and the thermal domain, respectively. These vectors may be considered as either hand-crafted features (e.g., DSIFT) or features from trainable neural networks.  Let $\mathbf{v_{i} \in V}$ and $\mathbf{t_{i} \in T}$, where the index $i$ denotes a pair of images $\mathbf{p}_i=(\mathbf{v_{i}},\mathbf{t_{i}})$ corresponding to the same subject. Without loss of generality, the pairs need not be precisely co-registered or synchronized.

Given a set of training pairs 
with known identities (i.e., labels), $y$,  our goal is to find a mapping $\mathbf{f}_{t} : \mathbf{V} \to \mathbf{T}$ (or $\mathbf{f}_{v} : \mathbf{T} \to \mathbf{V}$) such that:  $\mathbf{f}_{t}(\mathbf{v}_{i}) \approx \mathbf{t}_{i}$ (or $\mathbf{f}_{v}(\mathbf{t}_{i}) \approx \mathbf{v}_{i}$).  In either scenario, the visible representations are used as the reference in the gallery and the thermal representations are the test samples, which means both forms may be considered  thermal-to-visible FR. The key difference is whether the mapping is applied during enrollment or matching.

During deployment (i.e., testing), only the identities corresponding to the visible representations of enrolled subjects (i.e., gallery) are known.  Thus, we aim to learn a generalizable mapping $\mathbf{f}_t$ (or $\mathbf{f}_v$) that optimally discriminates genuine pairs and imposters pairs for the purposes of thermal-to-visible face identification. 

The problem with thermal-to-thermal (or visible-to-visible) FR is that the features, $\mathbf{t}$ (or $\mathbf{v}$), and classifiers, $\mathbf{\hat{y}}(\mathbf{t}; \mathbf{\Theta}_t)$ (or $\mathbf{\hat{y}}(\mathbf{v}; \mathbf{\Theta}_v)$), rely on distinct phenomenology. 
The within-spectrum classifiers are learned by minimizing the cross-entropy,
\begin{equation}
    \mathcal{L}(\mathbf{\Theta}_x)=-\sum \mathbf{y}\log\mathbf{\hat{y}}(\mathbf{x}; \mathbf{\Theta}_x),
    \label{eq:visible_cross_entropy}
\end{equation} 
between labels and predictions.  In Eq.~\ref{eq:visible_cross_entropy}, $\mathbf{x}$ denotes either $\mathbf{t}$ or $\mathbf{v}$, depending on the scenario.
Therefore, in principle, we also want to learn visible features that are optimally discriminative when fed into a thermal classifier (or vice versa).  So, in addition to Eq.~\ref{eq:visible_cross_entropy}, we also want to minimize
\begin{equation}
    \mathcal{L}(\mathbf{\Phi}_{x'})=-\sum \mathbf{y}\log\mathbf{\hat{y}}(\mathbf{f}_{x}(\mathbf{x}'; \Phi_{x'}); \mathbf{\Theta}_x),
    \label{eq:thermal_cross_entropy}
\end{equation}
where $\mathbf{x}'$ denotes features from the opposite domain of $\mathbf{x}$ (e.g., if $\mathbf{x}=\mathbf{t}$ then $\mathbf{x}'=\mathbf{v}$). $\mathbf{\Phi}_{x'}$ denotes the trainable parameters for the mapping $\mathbf{f}_x(\cdot)$.  

In following sections, we assume that $\mathbf{x}=\mathbf{t}$ then $\mathbf{x}'=\mathbf{v}$ to describe our approach.

\section{Proposed Domain Adaptation Framework}
\label{sec:method}
Our proposed domain adaptation framework (Figure~\ref{fig:model}) is composed of four main components:
\begin{compactitem}
    \item feature extraction using truncated deep neural network for both visible and thermal imagery (Section~\ref{ssec:feature}), 
    \item our proposed Residual Spectral Transform (RST) that bridges the remaining gap between visible and thermal features (Section~\ref{ssec:rst}),
    \item our proposed cross-domain identification loss that enhances holistic discriminability, especially when matching visible and thermal image representations (Section~\ref{ssec:idloss}),
    \item our proposed domain invariance loss that discourages domain predictability from visible or thermal features (Section~\ref{ssec:diloss}).
\end{compactitem}
Each of these components are discussed in depth in the following subsections.

\subsection{Feature Extraction}
\label{ssec:feature}
Feature representations, $\mathbf{v}$ and $\mathbf{t}$, are initially extracted from images using pre-trained neural networks, such as VGG16 and Resnet50 architecture.  However, since high-level features tend to be less transferable (i.e., high frequency details in visible spectrum facial imagery are lacking in the inherently smoother thermal facial imagery), we intentionally truncate the networks at the optimal depth in order to simultaneously obtain the largest receptive field and the most similar feature response.  

For the VGG16 network, we determined experimentally that the output of the third convolution-pooling block provided the most discriminative visible and thermal information for thermal-to-visible FR. 
Similarly, we found that the output of third residual block for the Resnet50 architecture provided the most robust thermal-to-visible FR performance.  Interestingly, the layers in which we truncate both of these networks result in feature maps with spatial dimensions of $25 \times 25$ ($H\times W$).  The feature maps do, however, have different number of channels ($C$);  VGG16 has 256 channels and ResNet50 has 512 channels.  Similar to \cite{DBLP:conf/bmvc/SarfrazS15}, we include a compression layer that is shared between visible and thermal representations.  This not only reduces the number of parameters in the subsequent layers, but also helps to eliminate factors associated with noise. 
The complete experiment details and results are provided in section~\ref{sec:ablation1}.

\subsection{Residual Spectral Transform}
\label{ssec:rst}
After extracting the most similar features possible, there is still a significant domain gap between the visible and thermal representations.  Therefore, we augment the visible (or thermal) network with our proposed Residual Spectral Transform (RST), which is the sub-network shown in Figure~\ref{fig:model}.

The RST is a residual block that aims to preserve as much discriminabilty from the truncated networks as possible while transforming features between thermal and visible domains. This sub-network transforms the features using three 1x1 convolutional layers: 
\begin{equation}
\mathcal{F}(\mathbf{u})= \text{Conv}_{c} \circ \text{Conv}_{200} \circ \text{Conv}_{200}(\mathbf{u}),
\end{equation}
where $\text{Conv}_k(\cdot)$ denotes the 1x1 convolutional layers with $k$ units and uses hyperbolic tangent activation function, and $c$ denotes the number of channels which depends on the selected network architecture or features.  Then, the RST is given by
\begin{equation}
    RST(\mathbf{u}) = \mathcal{F}(\mathbf{u}) + \mathbf{u}.
\end{equation}

\subsection{Cross-domain Identification Loss}
\label{ssec:idloss}
To preserve the identity of the images during FR, we use a cross-domain identification loss which is based on combining ~\eqref{eq:visible_cross_entropy} and ~\eqref{eq:thermal_cross_entropy}.   Therefore, our proposed loss function aims to simultaneously learn a discriminative thermal network (features and classifier) and optimal visible-to-thermal RST by minimizing 
\begin{equation}
    \mathcal{L}_{xID} =-\sum \left [\mathbf{y}\log\left \{\hat{\mathbf{y}}(\mathbf{t}; \mathbf{\Theta}_t ) \cdot \hat{\mathbf{y}}(\hat{\mathbf{t}}; \mathbf{\Theta}_t )\right \} \right ],
\label{eq:cross_domain_loss}
\end{equation}
where $\hat{\mathbf{t}}=\mathbf{f}_t(\mathbf{v}; \Phi_v)$.

Eq.~\ref{eq:cross_domain_loss} is minimized by alternately optimizing parameters $\mathbf{\Theta}_t$ and $\mathbf{\Phi}_v$, which learns discriminative information for the thermal domain while also performing cross-domain identification. 

\subsection{Domain Invariance Loss}
\label{ssec:diloss}
While Eq.~\ref{eq:cross_domain_loss} provides discriminative information, it is still possible to over-fit, resulting in inconsistent performance between training and inference.  In order to provide some regularization for better stability, we add a parallel domain classifier that estimates the probabilities that a given image representation comes from a visible image or a thermal image.  However, since our ultimate goal is to bridge the representational gap between visible and thermal imagery, optimal domain discriminability is not desirable because this would imply that there is still a detectable difference between visible and thermal feature representations.  Therefore, we add a regularizing loss function that encourages gradient-driven parameter updates to maintain poor domain predictability. 

Let a domain detector be denoted as $\mathcal{D}(\cdot)$, which produces two probabilities, $P(vis|\mathbf{h})$ and $P(thm|\mathbf{h})$, which are the probabilities that the feature representation $\mathbf{h}$ is computed from a visible ($vis$) or thermal ($thm$) image, respectively. 
Intuitively, we prefer to learn visible and thermal image representations such that $\mathcal{D}$ produces $P(vis|\mathbf{h}) \approx P(thm|\mathbf{h})$ for any $\mathbf{h} \in \{\mathbf{T},\mathbf{V}\}$.  Therefore, we add the following loss function to our proposed domain adaptation framework:
\begin{equation}
    \mathcal{L}_{\mathcal{D}} = -\sum \alpha\left \{\log \mathcal{D}(\mathbf{t}) + \log\mathcal{D}(\hat{\mathbf{t}}) \right \},
    \label{eq:domain_invar_loss}
\end{equation}
where $\alpha$ denotes the target probabilities.  In this case, $\alpha \in \mathbb{R}^2$ should be an approximately uniform distribution (i.e., all elements are close to $0.5$), which encourages poor domain prediction.

Therefore, the total loss function combines Eqs.~\ref{eq:cross_domain_loss} and \ref{eq:domain_invar_loss}, which yields 
\begin{equation}
    \mathcal{L}_{total}= (1-\lambda) \mathcal{L}_{xID} + \lambda \mathcal{L}_{\mathcal{D}},
    \label{eq:total_loss}
\end{equation}

\begin{table*}[htb]
    \centering
    \caption{Feature ablation study on protocol 1 showing Rank-1 identification (ID) rate and feature map dimensions (Dims.) }
    \label{tab:ablation1}
    \begin{tabular}{|c|c|c|}
        \hline
        \textbf{Method} & \textbf{Rank-1 ID} (\%) & \textbf{Feature Map Dims.} ($H\times W\times C$) \\
        \hline
        DOG+image\_patch & 2.83 & $24\times24\times400$\\
        DSIFT & 8.33 & $24\times24\times128$\\
        DOG+DSIFT & \textbf{10.50} & $\mathbf{24\times24\times128}$\\ \hline 
        DOG+VGG16(block5) + Avg. Pool & 12.67 & $1\times1\times512$\\
        DOG+VGG16(block5) & 15.17 & $6\times6\times512$\\
        DOG+VGG16(block4) & 46.67 & $12\times12\times512$\\
        DOG+VGG16(block3) & \textbf{58.83} & $\mathbf{25\times25\times256}$\\
        DOG+VGG16(block2) & 23.17 & $50\times50\times128$\\ \hline
        DOG+Resnet50 (5c) + Avg. Pool & 9.00 & $1\times1\times2048$\\
        DOG+Resnet50 (5c) & 10.00 & $7\times7\times2048$\\
        DOG+Resnet50 (4f) & 19.50 & $13\times13\times1024$\\
        DOG+Resnet50 (3d) & \textbf{70.83} & $\mathbf{25\times25\times512}$\\
        DOG+Resnet50 (2c) & 33.17 & $50\times50\times256$\\ \hline
        
    \end{tabular}
    
\end{table*}
 
\subsection{Implementation Details}
Pre-processing for thermal-to-visible FR applications typically includes image registration of corresponding visible and thermal images, image filtering, and cropping.
Using fiducial landmarks, including the center of eyes, base of nose, and mouth corners (i.e., 5-point alignment), the facial images are aligned to canonical coordinates using a similarity transform.  In practice, the landmarks can be automatically detected.  However, the datasets (section~\ref{sec:results}) used in this study include manually annotated landmarks for every image.



Next, a Difference of Gaussians (DoG) filter is applied to the registered images (visible and thermal). DoG filtering enhances the common edges between visible and thermal imagery, and has been shown to help for thermal-to-visible matching \cite{Hu:15}.

After filtering, all the images are cropped around the eyes, nose, and mouth to a $200\times200$ pixel image, in order to be consistent with past studies \cite{7477447} and to limit computational and memory requirements.  Additionally, this ``tight'' crop ensures that the network is attributing the identification with facial features rather than factors that may change frequently over time, such as hair style.  ``Better'' performance on benchmarks may be achieved when using a larger crop. However, larger crops may under-perform when significant variations, such as changes in hair style or image background, are observed.

When training our framework---the truncated VGG16 and Resnet50 architectures with our proposed RST---using the proposed cross-domain identification and domain invariance loss functions, $\lambda$ is set to 0.25.  Also, we used a compression ratio of 50\%---meaning the dimensionality is reduced by half.  Our compression ratio is consistent with the number of principal components used with DPM \cite{DBLP:conf/bmvc/SarfrazS15, DBLP:journals/ijcv/SarfrazS17}.
For reproducibility, our code can be found on \url{https://git.unl.edu/ece-unl-images-lab/cross-domain-identification}.

\section{Experiments \& Results}
\label{sec:results}
In this section, we first describe the protocols for our experimental analysis.  Then, we describe an ablation study used to determine how to best truncate the feature extraction models for thermal-to-visible FR. Next, we provide comprehensive analysis using two different protocols, including new analysis that examines non-frontal to frontal matching for thermal-to-visible FR. Lastly, we perform a second ablation study that examines the impact with and without the proposed domain invariance loss function.    

\begin{figure*}
\subfloat[]{\includegraphics[width=0.49\textwidth]{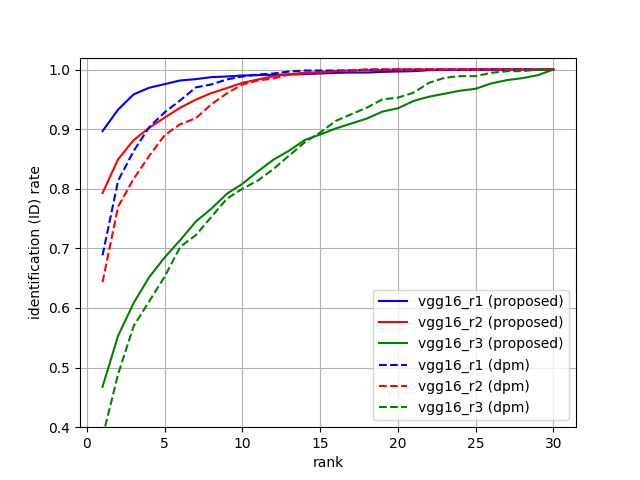}}
\subfloat[]{\includegraphics[width=0.49\textwidth]{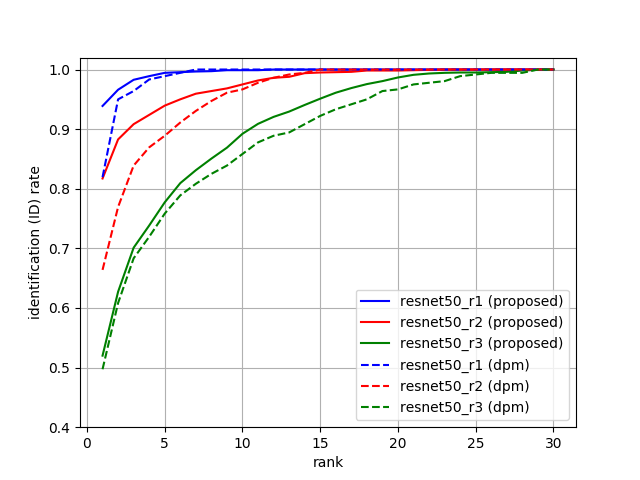}}
\label{fig:CMCplots}
\end{figure*}


\subsection{Protocols}
For experimental analysis, we use three separate datasets/protocols compiled by the CCDC Army Research Laboratory.  

The first dataset \cite{7789520}, referred to as ``protocol 1'' in \cite{article}, contains 2880 corresponding visible and polarimetric thermal image pairs from 60 unique subjects.  This collection contains imagery acquired with neutral expressions (baseline) and varying expressions at multiple standoff distances: $2.5m$ (r1), $5.0m$ (r2), and $7.5m$ (r3).  Using protocol 1, we train our proposed domain adaptation framework using only imagery at r1 (without any data augmentation) from 30 different subjects, and we evaluate using a 30 subject gallery with 4 baseline visible images per subjects (consistent with \cite{7789520, 7477447}).

The second dataset, referred to as ``protocol 2'' in \cite{article}, augments protocol 1 with an additional 1581 visible and polarimetric thermal image pairs from 51 subjects.  Therefore, protocol 2 contains 4461 image pairs from 111 subjects.  The additional 51 subject collection includes both baseline and varying expressions, but only acquired imagery at a single range equivalent to r1.
Complete details regarding this collection, including how to obtain the data, can be found in \cite{ 7789520, 7477447}.   
Using protocol 2, we train our proposed domain adaptation framework using only imagery at r1 (without any data augmentation) from 81 different subjects, and we evaluate using a 30 subject gallery with 4 baseline visible images per subjects (consistent with \cite{7789520, 7477447}).  

The third dataset, which we refer to as ``protocol 3,'' contains visible and thermal imagery from a separate collection (or volume) of 126 subjects.  There are a total of 5176 images (2198 visible and 2978 thermal) available from a single range consistent with r1.  However, unlike protocol 2, this collection contains faces with neutral expressions (baseline), varying expression, and varying pose (yaw in the range of $\pm60^\circ$). 
Using protocol 3, we train our proposed domain adaptation framework (without any data augmentation) using imagery from 96 different subjects, and we evaluate using a 30 subject gallery with 4 frontal, visible images per subject.  We intentionally use the same size gallery as protocol 2 for comparability.


\subsection{The Feature Ablation Study}
\label{sec:ablation1}
In this section, using protocol 1, we analyze the impact on thermal-to-visible FR for different features: image patches, DSIFT, VGG16, and Resnet50.  For VGG16 and Resnet50, we explore extracting features at various layers to determine optimal feature map size for the respective architectures. All features are extracted from the $200\times200$ visible and thermal images and compressed using principal component analysis (PCA) to 64 principal components.

The VGG16 architecture is composed of five convolution-pooling blocks (block1, $\dots$, block5) and several fully connected layers.  The Resnet50 architecture is composed of one convolultion layer and four residual blocks, and each residual block is composed of multiple residual layers.  For example, residual block 2 contains three residual layers (2a, 2b, 2c), residual block 3 contains four residual layers (3a, 3b, 3c, 3d), residual block 4 contains six residual layers (4a, 4b, 4c, 4d, 4e, 4f), and residual block 5 contains three residual layers (5a, 5b, 5c).  The final output (prior to the classifier) is average pooling layer that reduces the spatial dimensions of the features maps to $1\times1$.

Table~\ref{tab:ablation1} shows the rank-1 identification (ID) rate when matching thermal probe image representations with visible gallery image representations (from protocol 1) and the feature map dimensions (prior to compression). These image representations are matched using the cosine similarity measure. 
This table shows that better thermal-to-visible FR is achieved when extracting features from intermediate layers of the pre-trained VGG16 and Resnet50 networks, i.e., block3 and 3d, respectively.  This appears to indicate that extracting high-level features from the pre-trained networks depends too much on visible texture to perform well for cross-domain matching.  Also, lower level features have too small of a receptive field and insufficient context to perform cross-domain matching.  Given the aligned and cropped images, the best network performances seem to be attained when the feature maps' spatial dimensions are approximately $25\times25$.


\subsection{Proposed Framework Versus DPM}
Using both truncated VGG16 and Resnet50 architectures (Section~\ref{sec:ablation1}), we compared our proposed RST method (trained using Eq.~\ref{eq:total_loss}) with DPM (trained using Eq.~\ref{eq:dpm}).  Figure~\ref{fig:CMCplots} shows the cumulative match characteristic (CMC) curves for protocol 2.  First, note that our proposed domain adaptation framework exceeds the performance of DPM across all ranges and conditions when using both the truncated VGG16 and Resnet50 networks. Secondly, it is important to note that DSIFT + DPM \cite{7477447} achieves 84\%, 75\%, 58\% performance for r1, r2, and r3 and our proposed framework achieves 94.2\%, 81.7\%, and 52.0\%.  Thus, our approach improves performance for r1 and r2, but r3 performance is somewhat lacking still.  This may be due to either a trade-off between DSIFT and neural networks based features (which can be more sensitive to low-quality images) or the fact that we only train using r1 imagery.  Thus, it may be possible to see gains by incorporating multiple resolutions or data augmentation during training in order to boost the r3 performance. 

We also compared the DPM and our proposed domain adaptation framework on protocol 3, which includes more subject and more variations (particularly pose).  Table~\ref{tab:protocol3} shows the rank-1 ID for the truncated VGG16 and Resnet50 architectures under both variable expression and pose conditions.  It is important to note that the gallery is composed of only frontal imagery.  Unsurprisingly, the performance drops when matching non-frontal thermal imagery to the frontal visible gallery.  However, our proposed framework still achieves better rank-1 performance than DPM.

\begin{table}[htb]
\caption{Protocol 3 Rank-1 ID performance for pose and expression variations}
\label{tab:protocol3}
\begin{tabular}{ |p{1.5cm}|p{3cm}|p{2.3cm}| }
\hline
\textbf{Condition} & \textbf{Method} & \textbf{Rank-1 ID (\%)} \\
\hline
\multirow{4}{*}{Expression} & Resnet50 + DPM & 91.56 \\ 
& VGG16 + DPM & 82.29 \\
&  Resnet50 + Proposed & \textbf{96.00} \\ 
& VGG16 + Proposed & \textbf{84.00} \\ 
\hline
\multirow{4}{*}{Pose} & Resnet50 + DPM & 24.42 \\ 
& VGG16 + DPM & 21.33 \\ 
& Resnet50 + Proposed & \textbf{29.91} \\ 
& VGG16 + Proposed & \textbf{21.38} \\ 
\hline
\end{tabular}
\end{table}

\subsection{Effect of Domain Invariance Loss}
This ablation study aims to empirically assess the effect of the domain invariance loss function (Eq.~\ref{eq:domain_invar_loss}). Using protocol 2, we consider our proposed model for scenario 1---finding a mapping $\mathbf{f}_{t} : \mathbf{V} \to \mathbf{T}$---and scenario 2---finding a mapping $\mathbf{f}_{v} : \mathbf{T} \to \mathbf{V}$.  In each scenario, we compare the rank-1 performance at r1 both with ($\lambda=0.25$) and without ($\lambda=0$) the domain invariance loss.  Additionally, we also consider both the truncated VGG16 and Resnet50 models.

Table~\ref{tab:ablation2} shows that using VGG16 with the domain invariance loss improves performance by 17.50\% and 31.33\% under scenario 1 and scenario 2, respectively.  Also, the results show that using Resnet50 with the domain invariance loss improves performance by 5.84\% and 1.66\% under scenario 1 and scenario 2, respectively. This ablation study demonstrates that the domain invariance loss is an important aspect of the proposed framework.      

\begin{table}[htb]
\caption{Effect of domain invariance loss for scenario 1 and scenario 2}
\label{tab:ablation2}
\centering
\begin{tabular}{ |c|c|c|c| } 
\hline
\textbf{Scenario} & \textbf{Domain Invariance} & \textbf{Resnet50} & \textbf{VGG16} \\ 
\hline
\multirow{2}{*}{Scenario 1} & no & 88.33\% & 65.83\% \\
& yes & 94.17\% & 83.33\%  \\
\hline
\multirow{2}{*}{Scenario 2} & no & 89.17\% & 49.17\%  \\
& yes & 90.83\% & 80.50\%  \\
\hline
\end{tabular}
\end{table}

\section{Conclusion}
In this paper, we proposed a new domain adaptation framework, which used truncated deep neural network with a new RST sub-network.  This framework was trained using our new cross-domain identification and domain invariance loss function.  Compared with DPM, which is trained using a Euclidean loss function between thermal and visible features, our framework shows significant improvements across multiple ranges, poses, and expressions.  Also, we demonstrated significant improvement to state-of-the-art features extraction methods, like DSIFT and DPM, and show that our domain invariance loss function plays an important role in achieving robust thermal-to-visible FR.

Most importantly, we have developed a framework that alleviates the restrictive need for precisely registered and synchronized imagery because we introduce two loss functions that operated at the task-level rather than the feature-level.  We hypothesize with larger datasets (on the order of 300+ subjects and 500,000+ thermal-visible pairs) will provide further generalization to the challenging non-frontal to frontal matching for thermal-to-visible FR.


\ifijcbfinal
\section{Acknowledgments}
This research was partially supported by Booz Allen Hamilton (BAH) and the U.S. Army Combat Capabilities Development Command (CCDC) Army Research Laboratory (ARL) under contract W911QX-17-D-0015. The views and conclusions contained in this document are those of the authors and should not be interpreted as representing the official policies, either expressed or implied, of the BAH, CCDC ARL, or the U.S. Government. The U.S. Government is authorized to reproduce and distribute reprints for Government purposes notwithstanding any copyright notation herein.
\fi

{\small
\bibliographystyle{ieee}
\bibliography{main}
}
\end{document}